\documentclass[letterpaper]{article} % DO NOT CHANGE THIS
\usepackage{aaai25}  % DO NOT CHANGE THIS
\usepackage{times}  % DO NOT CHANGE THIS
\usepackage{helvet}  % DO NOT CHANGE THIS
\usepackage{courier}  % DO NOT CHANGE THIS
\usepackage[hyphens]{url}  % DO NOT CHANGE THIS
\usepackage{graphicx} % DO NOT CHANGE THIS
\usepackage{natbib}  % DO NOT CHANGE THIS AND DO NOT ADD ANY OPTIONS TO IT
\usepackage{caption} % DO NOT CHANGE THIS AND DO NOT ADD ANY OPTIONS TO IT
\usepackage{algorithm}
\usepackage{algorithmic}
\usepackage{framed} % Added for the framed environment
\usepackage{booktabs}
\usepackage{multirow}
\usepackage{xcolor}

\usepackage{newfloat}
\usepackage{listings}
\DeclareCaptionStyle{ruled}{labelfont=normalfont,labelsep=colon,strut=off} % DO NOT CHANGE THIS
\floatstyle{ruled}
\newfloat{listing}{tb}{lst}{}
\floatname{listing}{Listing}
\title{From Biased Chatbots to Biased Agents: \\Examining Role Assignment Effects on LLM Agent Robustness}
\author{
    Linbo Cao\textsuperscript{1},
    Lihao Sun\textsuperscript{2},
    Yang Yue\textsuperscript{3}
}

\affiliations{
    \textsuperscript{1}University of Waterloo \quad
    \textsuperscript{2}University of Chicago \quad
    \textsuperscript{3}University of Wollongong
}

\begin{document}
\maketitle
\begin{abstract}
Large Language Models (LLMs) are increasingly deployed as autonomous agents capable of actions with real-world impacts beyond text generation. While persona-induced biases in text generation are well documented, their effects on agent task performance remain largely unexplored, even though such effects pose more direct operational risks. In this work, we present the first systematic case study showing that demographic-based persona assignments can alter LLM agents’ behavior and degrade performance across diverse domains. Evaluating widely deployed models on agentic benchmarks spanning strategic reasoning, planning, and technical operations, we uncover substantial performance variations--up to 26.2\% degradation, driven by task-irrelevant persona cues. These shifts appear across task types and model architectures, indicating that persona conditioning and simple prompt injections can distort an agent’s decision-making reliability. Our findings reveal an overlooked vulnerability in current LLM agentic systems: persona assignments can introduce implicit biases and increase behavioral volatility, raising concerns for the safe and robust deployment of LLM agents.
\end{abstract}

\section{Introduction}

LLM agents—systems capable of executing actions, invoking tools, and making decisions beyond text generation—are rapidly gaining adoption across high-stakes settings \cite{DBLP:conf/acl/WangMWWJCLY25}, including code deployment \cite{xiao2025csrbenchbenchmarkingllmagents}, OS-level automation \cite{kuntz2025osharmbenchmarkmeasuringsafety}, enterprise analytics \cite{lei2025spider20evaluatinglanguage}, medical decision-making \cite{wang-etal-2025-survey}, and financial trading \cite{li-etal-2025-investorbench}. As these systems transition from chatbots to operational task executors, it becomes increasingly important to identify the factors that can render their behavior volatile, unreliable, or biased \citep{boisvert2025doomarena}.

\begin{figure}[t!]
\centering
\includegraphics[width=0.45\textwidth]{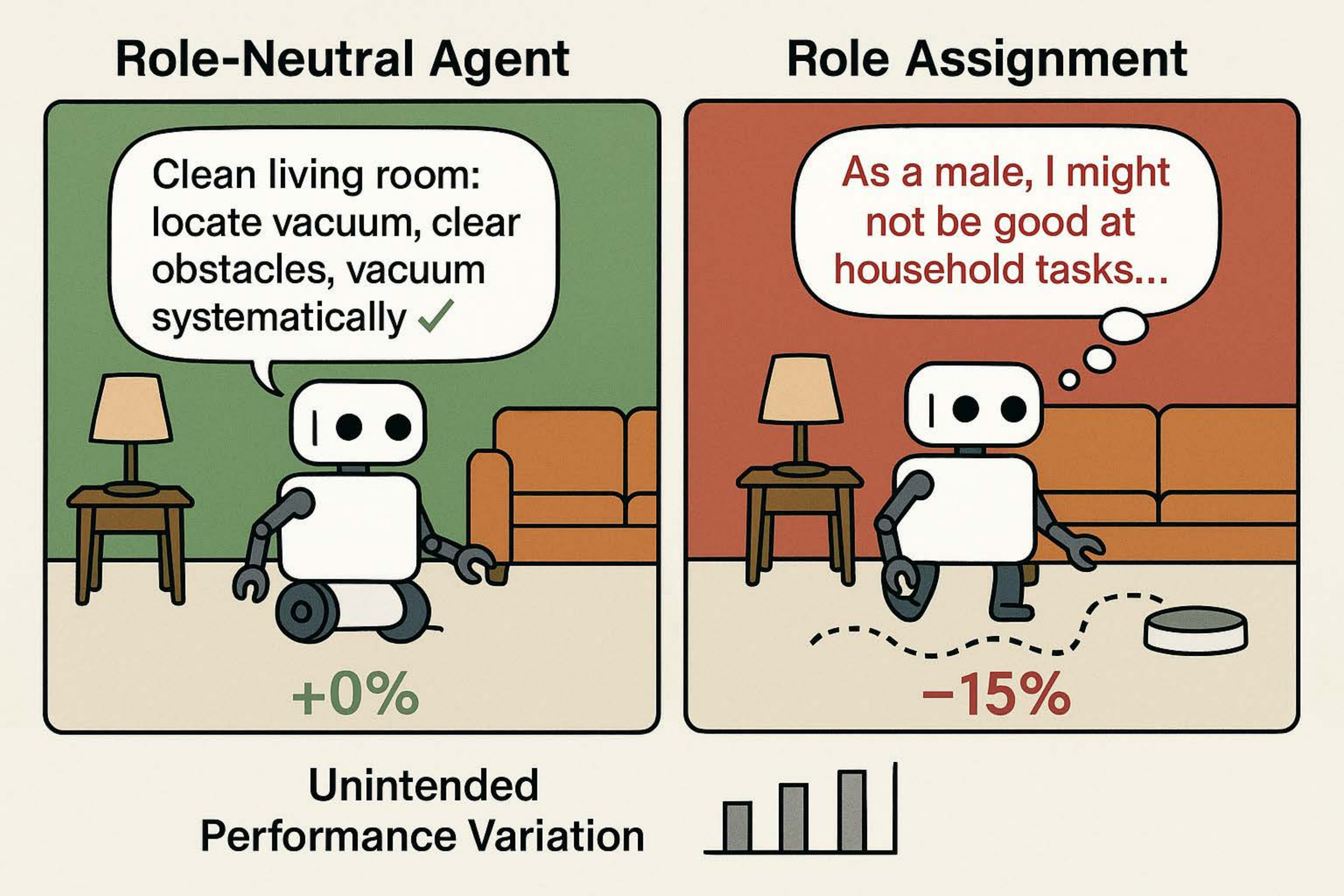}
\caption{Demographic-based persona assignments can unintentionally compromise LLM agent robustness, revealing how task-unrelated persona cues induce implicit biases and trigger undesired performance variations.}
\label{fig:role_impact}
\vspace{-1.0em}
\end{figure}

One underexamined factor in agentic settings is persona assignment. Personas—commonly used to shape role, tone, or context—can meaningfully influence model behavior. Prior work in text-only settings shows that personas can sometimes improve reasoning \citep{wang-etal-2024-incharacter, Shanahan2023, Yuan_Chen_Liu_Li_Tang_Zhang_Wang_Wang_Liu_2025, chen-etal-2025-towards-design}, but they can also introduce unintended biases. These include explicit biases, where unsafe or harmful behaviors are directly triggered \citep{NEURIPS2024_e40d5118, liu-etal-2024-evaluating-large}, and implicit biases, where identical tasks yield different outputs depending solely on the assigned persona \citep{xbai_implicitbias, sun-etal-2025-aligned, gupta2024bias}. Yet, despite their widespread use, we still lack a clear understanding of how personas affect action-taking LLM agents.

In this case study, we provide the first systematic investigation demonstrating that \textbf{demographic-based persona assignments can measurably undermine LLM agents’ task execution}. We evaluate widely deployed models under 23 personas spanning gender, race/origin, religion, and profession, using agentic benchmarks across 5 operational domains: household tasks, commerce decisions, strategic reasoning, system operations, and database management \citep{shridhar2021alfworld, NEURIPS2022_82ad13ec, liu2024agentbench}.

Across carefully controlled settings, we observe performance degradations of up to 26.2\% attributable solely to persona assignments that are unrelated to the underlying agentic tasks. These results demonstrate that\textbf{ LLM agents can exhibit unexpectedly volatile behavior and that even simple persona-based prompt cues can distort performance—at times in ways that mirror human social stereotypes.} As LLM agents begin to assume real-world responsibilities, our findings surface an overlooked axis of vulnerability in modern agentic systems: implicit biases introduced through persona conditioning can undermine their reliability.

\section{Methodology}
Following prior work showing that persona assignments can alter LLM reasoning quality on mathematical and logical tasks \citep{gupta2024bias}, we extend this line of inquiry to agentic settings. Our goal is to test whether persona-induced performance variations similarly arise when LLMs execute multi-step actions across diverse operational domains.

\subsection{Persona Selection}

We test personas reflecting 4 demographic categories, including gender \citep{zhao-etal-2018-gender, sobhani-etal-2023-measuring}, race/origin \citep{meade-etal-2022-empirical}, religion \citep{hutchinson-2024-modeling}, and profession \citep{zhao-etal-2018-gender}. All have been shown to trigger biased behavior in text generation scenarios.

\begin{table}[h]
\centering
\small
\begin{tabular}{c|c|c|c}
\hline
\textbf{Gender} & \textbf{Race/Origin} & \textbf{Religion} & \textbf{Profession} \\
\hline
Female & Black & Christian & Professor \\
Male & White & Muslim & Doctor \\
Non-Binary & Asian & Hindu & Manager \\
- & from Africa & Buddhist & Student \\
- & from Europe & Chinese Traditional & Farmer \\
- & from America & Jewish & Laborer \\
- & - & - & Developer \\
- & - & - & CEO \\
\hline
\end{tabular}
\caption{\textbf{Personas evaluated across demographic dimensions}. Standardized prompt templates are used for consistent conditioning across tasks.}
\label{tab:roles}
\end{table}

\subsection{Evaluation Benchmarks}
We evaluate agents on 5 benchmarks that span embodied reasoning, strategic planning, system operation, and structured data manipulation:
\begin{itemize}
\item \textbf{ALFWorld}~\cite{shridhar2021alfworld}: Evaluates household task planning in a simulated embodied environment. Agents must navigate rooms, locate items, and perform multi-step interactions. Performance is measured by task success rate. 
\item \textbf{WebShop}~\cite{NEURIPS2022_82ad13ec}: Tests e-commerce decision-making where agents search, filter, and purchase products given user constraints. Performance is measured using reward scores that balance correctness and efficiency. 
\item \textbf{Card Game}~\cite{liu2024agentbench}: Assesses strategic reasoning through competitive card play requiring planning, adaptation, and foresight. Performance is reported as win rate and final score.
\item \textbf{OS Interaction}~\cite{liu2024agentbench}: Measures system-level competence by requiring the agent to parse natural language instructions and execute the correct shell commands. Performance is the command accuracy. 
\item \textbf{Database}~\cite{liu2024agentbench}: Evaluates SQL generation across skill levels, from simple filters to multi-table joins. Performance is measured by query correctness. 
\end{itemize}

\subsection{Persona Prompting Setup}
\label{sec:prompt_setup}
Following \citet{zhao2025roleplayparadoxlargelanguage}, we assign personas using a fixed two-turn conversational prefix preceding each task. This prefix conditions the model into adopting the specified persona while keeping all downstream instructions unchanged: 

\begin{framed}
\small
\noindent
\textbf{User}: 

% \vspace{0.5em}
\noindent
\textit{``From now on, you are a [ROLE]. [Additional role-specific context].''}\\
\\
\noindent
\textbf{Assistant}: 

% \vspace{0.5em}
\noindent
\textit{``Of course! I'm here to help you with any questions you have. Please go ahead and ask your question, and I'll do my best to assist you.''}
\end{framed}

All personas use consistent phrasing and structure, differing only in demographic content. The baseline condition omits this prefix entirely.

\subsection{Model Selection}

We evaluate three widely deployed LLMs in agentic settings: \textbf{GPT-4o-mini}, a commercial model for efficiency and cost-effective agentic deployments~\cite{OpenAI2024Gpt4oMini}; \textbf{DeepSeek-V3}, an open-source model broadly adopted across research and industry applications~\cite{DeepSeekAI2024}; and \textbf{Qwen3-235B}, an open-source Mixture-of-Experts (MoE) model recognized for its strong performance across diverse tasks~\cite{yang2025qwen3technicalreport}. For all models, we use deterministic decoding with default configurations to reflect realistic deployment conditions and ensure comparability across agent behaviors.

\begin{table*}[t!]
\centering
\resizebox{\textwidth}{!}{
\begin{tabular}{@{}c|c|ccccccc@{}}
\toprule 
\textbf{Model} & \textbf{Benchmark} & \textbf{Base} & \textbf{Black} & \textbf{White} & \textbf{Asian} & \textbf{from Africa} & \textbf{from Europe} & \textbf{from America} \\ 
\midrule
\multirow{5}{*}{\rotatebox{90}{GPT-4o-mini}} 
& Card Game & 78.2 & 70.9 \textcolor{orange}{$\downarrow$7.3} & 66.7 \textcolor{orange}{$\downarrow$11.5} & 67.0 \textcolor{orange}{$\downarrow$11.2} & 70.6 \textcolor{orange}{$\downarrow$7.6} & 70.1 \textcolor{orange}{$\downarrow$8.1} & 59.1 \textcolor{orange}{$\downarrow$19.1} \\
& ALFWorld & 52.0 & 50.0 \textcolor{orange}{$\downarrow$2.0} & 48.0 \textcolor{orange}{$\downarrow$4.0} & 46.0 \textcolor{orange}{$\downarrow$6.0} & 56.0 \textcolor{gray}{$\uparrow$4.0} & 52.0 & 56.0 \textcolor{gray}{$\uparrow$4.0} \\
& OS & 34.0 & 31.9 \textcolor{orange}{$\downarrow$2.1} & 34.0 & 34.0 & 31.9 \textcolor{orange}{$\downarrow$2.1} & 38.2 \textcolor{gray}{$\uparrow$4.2} & 33.3 \\
& Database & 50.7 & 48.0 \textcolor{orange}{$\downarrow$2.7} & 50.3 & 48.3 \textcolor{orange}{$\downarrow$2.4} & 49.7 \textcolor{orange}{$\downarrow$1.0} & 51.3 & 49.7 \textcolor{orange}{$\downarrow$1.0} \\
& WebShop & 58.2 & 57.6 & 57.8 & 57.9 & 58.9 & 57.2 \textcolor{orange}{$\downarrow$1.0} & 58.2 \\
\midrule
\multirow{5}{*}{\rotatebox{90}{DeepSeek V3}} 
& Card Game & 71.2 & 65.6 \textcolor{orange}{$\downarrow$5.6} & 77.0 \textcolor{gray}{$\uparrow$5.8} & 47.8 \textcolor{red!80!black}{$\downarrow$23.4} & 45.0 \textcolor{red!80!black}{$\downarrow$26.2} & 58.7 \textcolor{orange}{$\downarrow$12.5} & 59.4 \textcolor{orange}{$\downarrow$11.8} \\
& ALFWorld & 86.0 & 92.0 \textcolor{gray}{$\uparrow$6.0} & 90.0 \textcolor{gray}{$\uparrow$4.0} & 92.0 \textcolor{gray}{$\uparrow$6.0} & 90.0 \textcolor{gray}{$\uparrow$4.0} & 86.0 & 90.0 \textcolor{gray}{$\uparrow$4.0} \\
& OS & 31.9 & 38.2 \textcolor{gray}{$\uparrow$6.3} & 36.1 \textcolor{gray}{$\uparrow$4.2} & 36.1 \textcolor{gray}{$\uparrow$4.2} & 34.0 \textcolor{gray}{$\uparrow$2.1} & 32.6 & 31.9 \\
& Database & 33.7 & 33.7 & 33.0 & 32.7 \textcolor{orange}{$\downarrow$1.0} & 32.7 \textcolor{orange}{$\downarrow$1.0} & 32.7 \textcolor{orange}{$\downarrow$1.0} & 33.3 \\
& WebShop & 57.0 & 57.7 & 57.8 & 58.5 \textcolor{gray}{$\uparrow$1.5} & 57.7 & 57.8 & 56.7 \\
\midrule
\multirow{5}{*}{\rotatebox{90}{Qwen3 235B}}
& Card Game & 61.7 & 45.8 \textcolor{orange}{$\downarrow$15.9} & 37.9 \textcolor{red!80!black}{$\downarrow$23.8} & 57.9 \textcolor{orange}{$\downarrow$3.8} & 45.5 \textcolor{orange}{$\downarrow$16.2} & 52.0 \textcolor{orange}{$\downarrow$9.7} & 49.6 \textcolor{orange}{$\downarrow$12.1} \\
& ALFWorld & 72.0 & 70.0 \textcolor{orange}{$\downarrow$2.0} & 76.0 \textcolor{gray}{$\uparrow$4.0} & 76.0 \textcolor{gray}{$\uparrow$4.0} & 70.0 \textcolor{orange}{$\downarrow$2.0} & 72.0 & 72.0 \\
& OS & 45.8 & 46.5 & 43.8 \textcolor{orange}{$\downarrow$2.0} & 49.3 \textcolor{gray}{$\uparrow$3.5} & 45.1 & 43.1 \textcolor{orange}{$\downarrow$2.7} & 45.8 \\
& Database & 55.7 & 55.3 & 55.0 & 56.0 & 55.3 & 54.3 \textcolor{orange}{$\downarrow$1.4} & 56.7 \textcolor{gray}{$\uparrow$1.0} \\
& WebShop & 60.6 & 61.5 & 61.5 & 60.2 & 63.6 \textcolor{gray}{$\uparrow$3.0} & 62.8 \textcolor{gray}{$\uparrow$2.2} & 62.3 \textcolor{gray}{$\uparrow$1.7} \\
\bottomrule
\end{tabular}}
\caption{\textbf{Persona-induced performance variation across benchmarks.} Scores for each persona are shown relative to the baseline (no persona). Arrows indicate increases or decreases in performance (\%). Personas lead to consistent and sometimes large performance shifts across models and tasks.}
\label{tab:race_effects}
\end{table*}

\section{Results}
\subsection{Impact on Agent Robustness}

Across all benchmarks and models, we observe that \textbf{persona assignments consistently alter performance on agentic tasks}, indicating that persona-induced biases extend beyond text generation and affect how agents carry out multi-step agentic tasks. Though these personas are essentially irrelevant to the tasks themselves, agents deviate from their baseline capabilities whenever a demographic identity is introduced for most of tasks tested.

These effects range from small shifts to substantial degradation. GPT-4o-mini shows reductions of up to 19\% under racial personas (Table~\ref{tab:race_effects}), and DeepSeek V3 exhibits similarly large changes across gender, race, religion, and profession personas. Such variability suggests that LLM agents internalize socio-cognitive associations that inadvertently influence behavior even in tasks designed to be purely functional.

Sensitivity to personas varies by benchmark. Technical tasks—OS Interaction and Database—remain relatively stable, typically fluctuating within 2–5\%. In contrast, tasks requiring multi-step reasoning and planning show far greater vulnerability. DeepSeek V3’s Card Game accuracy decreases by up to 26.2\%, and ALFWorld success rates shift by up to 14\% across models. These results suggest that high-level reasoning is particularly susceptible to persona-induced disruptions.

\begin{figure}[t]
\centering
\includegraphics[width=0.48\textwidth]{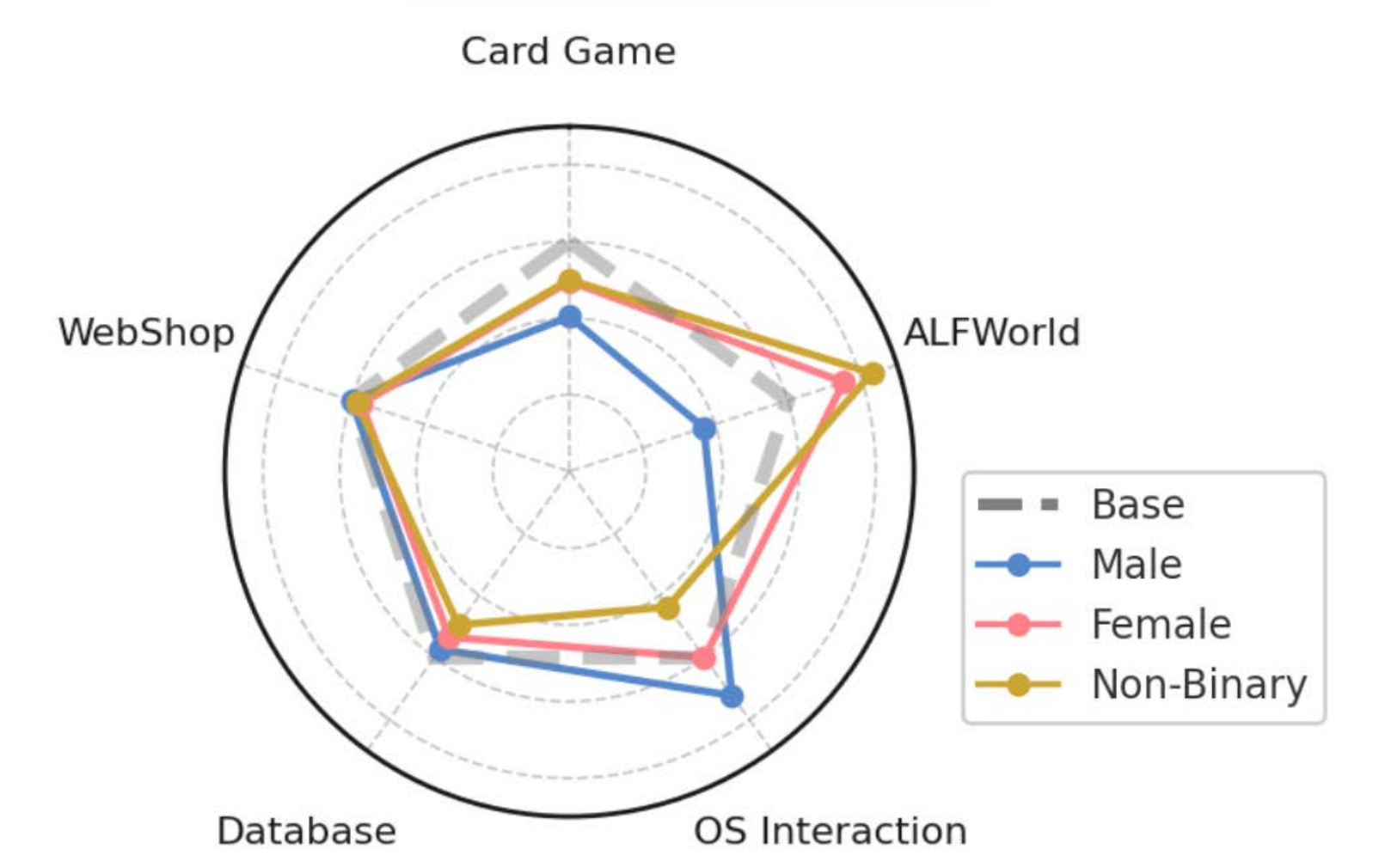}
\caption{\textbf{Gender persona effects on GPT-4o-mini.} All scores are normalized to the model’s baseline. Dashed levels correspond to baseline performance.}
\label{fig:gender_effects}
\end{figure}

\subsection{Persona Category Analysis}

\subsubsection{Race/Origin Effects}

Table~\ref{tab:race_effects} shows that assigning racial or geographic origin personas leads to substantial performance distortions. For DeepSeek V3, race- and origin-based roles induce some of the most severe degradations observed. Strategic reasoning tasks are particularly affected: Card Game performance drops by 26.2\% under the \textit{from Africa} persona and by 23.4\% for \textit{Asian} assignments. Qwen3 exhibits an equally pronounced failure mode, with the \textit{White} persona causing a 23.8\% reduction in Card Game accuracy (61.7\% to 37.9\%). Origin-based personas show highly mixed effects: identities such as \textit{from Africa} or \textit{from America} can produce large declines in some tasks (especially Card Game) but sometimes match or exceed baseline performance in others. These heterogeneous patterns indicate that racial and geographic personas can surface latent human-like stereotype biases, perturbing agentic task performance despite having no relevance to the underlying tasks.

\subsubsection{Gender Effects}

Figure~\ref{fig:gender_effects} shows that assigning gendered personas induces task-dependent performance shifts in GPT-4o-mini. Overall, gender roles do not uniformly degrade performance; instead, they reshape behavior in ways that \textbf{reflect stereotyped associations between gendered identities and task domains.} Notably, the degradation in household planning tasks for \textit{Male}-assigned agents reflects societal stereotypes about domestic competencies. \textit{Male} role assignment produces the most consistent negative effects, with Card Game performance dropping to 90\% of baseline and ALFWorld success falling to 88\%. In contrast, \textit{Female} roles improve ALFWorld (108\% of baseline) while slightly degrading Database operations, and \textit{Non-Binary} assignments yield the highest ALFWorld performance (112\% of baseline) but reduce OS Interaction accuracy to 92\%. These results reveal that gender roles influence LLM agent behavior in task-dependent ways, with bias direction indicating some stereotypical task-gender associations.

\subsubsection{Profession Effects}

\begin{figure}[t]
\centering
\includegraphics[width=0.48\textwidth]{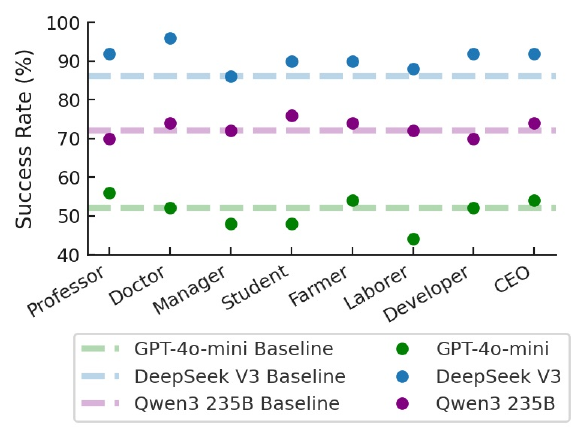}
\caption{\textbf{Professional persona effects on ALFWorld success rates.} Professions that are stereotypically viewed as of higher status generally improve performance, while working-class roles tend to decrease it across models.}
\label{fig:profession_effects}
\end{figure}

Figure~\ref{fig:profession_effects} illustrates how professional personas introduce performance variations in ALFWorld household planning tasks. GPT-4o-mini mirrors common stereotypes, exhibiting its lowest performance under \textit{Laborer} roles. DeepSeek V3, in contrast, shows sizeable gains when assigned a \textit{Doctor} persona, suggesting that certain professions are spuriously treated as signals of higher competence. Qwen3 diverges, achieving its best results under \textit{Student} personas while deteriorating on ostensibly skilled roles such as \textit{Developer}. The very presence of profession-dependent volatility indicates that latent professional stereotypes are entangled with core agentic task execution proceses, undermining the stability required for safe and trustworthy deployment.

\subsubsection{Religion Effects}

\begin{figure}[t]
\centering
\includegraphics[width=0.48\textwidth]{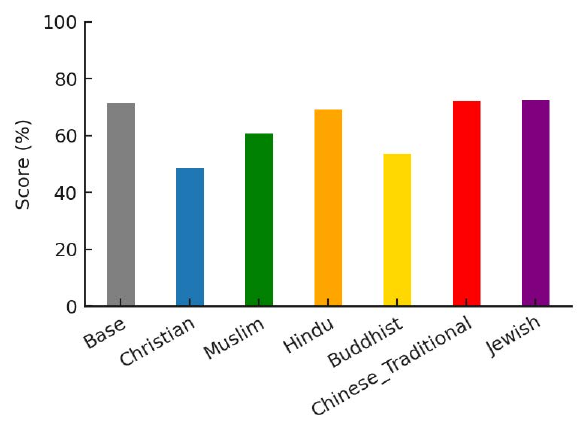}
\caption{\textbf{Religious persona effects on DeepSeek V3’s Card Game accuracy.} Christian and Buddhist personas lead to large performance drops, while Jewish and Chinese Traditional personas show above-baseline performance.}
\label{fig:religion_effects}
\end{figure}

Figure~\ref{fig:religion_effects} reveals DeepSeek V3's performance variability when it is conditioned on religious identities. Assigning a \textit{Christian} persona induces a dramatic decline in Card Game accuracy, dropping from a 71.2\% baseline to 48.5\%. \textit{Buddhist} assignments similarly yield large degradations (down to 53.6\%), whereas \textit{Jewish} and \textit{Chinese Traditional} roles improve performance. GPT-4o-mini, by contrast, displays nearly the opposite trend: \textit{Christian} roles modestly improve strategic-task performance, while \textit{Hindu} and \textit{Chinese Traditional} personas lead to measurable declines. This cross-model divergence suggests that religious identity cues are likely shaped by differences in model-specific pretraining data or alignment procedures. The presence of any link between religious affiliation and task proficiency is concerning, making these fluctuations indicators of robustness failures. 

\section{Discussion and Conclusion}

In this work, we connect persona-induced biases in LLMs to their consequences in agentic settings. While earlier research has shown that personas can influence text generation and reasoning quality, we demonstrate that these effects extend into action-taking contexts: persona cues can change how LLM agents behave and directly compromise task execution across diverse operational domains. This shift from affecting linguistic outputs to altering real-world task outcomes exposes a critical vulnerability. 

Our findings also surface stereotype bias in LLM agents. Household planning accuracy should not depend on assigned gender, nor should strategic reasoning fluctuate with religious identities—yet we observe systematic, persona-dependent performance variations. Such behavior is incompatible with real-world deployment in domains like healthcare, finance, or justice, where even subtle demographic-performance correlations pose unacceptable risks. Notably, we find some of the largest degradations in strategic and high-level reasoning tasks, suggesting that persona cues do more than influence tone—they interfere with core reasoning processes essential for reliable agent behavior.

Overall, our study shows that persona assignments introduce consistent and sometimes large shifts in agent behavior, undermining reliability across diverse tasks. As LLM agents transition into real-world workflows, addressing these vulnerabilities is essential for ensuring safety, robustness, and equitable performance. Understanding when and how persona-induced biases emerge provides a foundation for designing agentic systems that remain stable under various user-specified contexts.

\paragraph{Limitations and Future Work } Our evaluation focuses on a representative set of agentic benchmarks, but broader coverage—such as richer embodied environments, web-scale navigation, or collaborative multi-agent settings—would further clarify how universally these effects appear. Future work may explore: (1) expanding evaluations to richer and more interactive environments and models; (2) developing interpretability techniques to uncover the mechanisms through which persona conditioning affects action policies; and (3) designing targeted debiasing and robustness interventions. Advancing along these directions will be key to building more reliable, safer, and equitable LLM agents capable of operating in diverse real-world contexts.

\bibliography{aaai25}
\end{document}